\documentclass{article}

\PassOptionsToPackage{numbers, compress}{natbib}
\usepackage[preprint]{neurips_2023}
\usepackage{graphicx}
\usepackage{subcaption}

\usepackage[utf8]{inputenc} % allow utf-8 input
\usepackage[T1]{fontenc}    % use 8-bit T1 fonts
\usepackage{hyperref}       % hyperlinks
\usepackage{url}            % simple URL typesetting
\usepackage{booktabs}       % professional-quality tables
\usepackage{amsfonts}       % blackboard math symbols
\usepackage{nicefrac}       % compact symbols for 1/2, etc.
\usepackage{microtype}      % microtypography
\usepackage{xcolor}         % colors

\title{Contextual fusion enhances robustness to image blurring}

\author{
    Shruti Joshi\\
  Department of Medicine\\
  University of California, San Diego\\
  La Jolla, CA 92093\\
  \texttt{s4joshi@ucsd.edu} \\
    \And
  Aiswarya Akumalla \\
  Department of Medicine\\
  University of California, San Diego\\
  La Jolla, CA 92093\\
  \texttt{aakumall@eng.ucsd.edu} \\
    \AND
    Seth Haney \\
    Department of Medicine\\
  University of California, San Diego\\
  La Jolla, CA 92093\\
  \texttt{sethdhaney@gmail.com} \\
      \And
    Maxim Bazhenov \\
    Department of Medicine\\
  University of California, San Diego\\
  La Jolla, CA 92093\\
  \texttt{mbazhenov@ucsd.edu} 
}

\begin{document}

\maketitle

\begin{abstract}
Mammalian brains handle complex reasoning by integrating information across brain regions specialized for particular sensory modalities. This enables improved robustness and generalization versus deep neural networks, which typically process one modality and are vulnerable to perturbations. While defense methods exist, they do not generalize well across perturbations. We developed a fusion model combining background and foreground features from CNNs trained on Imagenet and Places365. We tested its robustness to human-perceivable perturbations on MS COCO. The fusion model improved robustness, especially for classes with greater context variability. Our proposed solution for integrating multiple modalities provides a new approach to enhance robustness and may be complementary to existing methods.
%or maintain clean data performance. 

\end{abstract}

\section{Introduction}
Current deep neural networks excel at solving specific tasks for certain input types (e.g., convolutional neural networks for images), but generally struggle to combine different input types (e.g., visual, semantic, auditory) into a unified representation. Some of the challenges include finding the right alignment of unimodal representations, fusion strategy, and complexity measures for determining the efficacy of fused representations \cite{DBLP:journals/corr/BaltrusaitisAM17}. In contrast, biological systems adeptly form coherent object representations and use them for various tasks by linking features from specialized cortical networks \cite{gisiger_computational_2000, pandya_association_1982,mars_412_2017,rosen_role_2017}. Thus, there is a need to develop approaches that would combine strength of specialized networks with ability to represent information across multiple streams as human and animal brain can do efficiently.

\section{Methodology}

\subsection{Multimodal model}

We developed a fusion-based multimodal model which combines the features extracted from the foreground and the background of an image to perform a classification task. The foreground and background feature extractors are based on the Resnet18 \cite{zhou_learning_2014} architecture but trained with different training data. The foreground one was trained on the Imagenet \cite{imagenet} dataset, whereas the background one was trained on Places365 \cite{zhou_learning_2014} dataset. Here, scene-based features provide context for the object and hence, we expect it to aid object classification.  %The Places365 dataset has 365 scene-centric image classes and the background feature extractor can encode scene-based features. 

\begin{figure}
  \centering
    \includegraphics[width=0.65\textwidth]{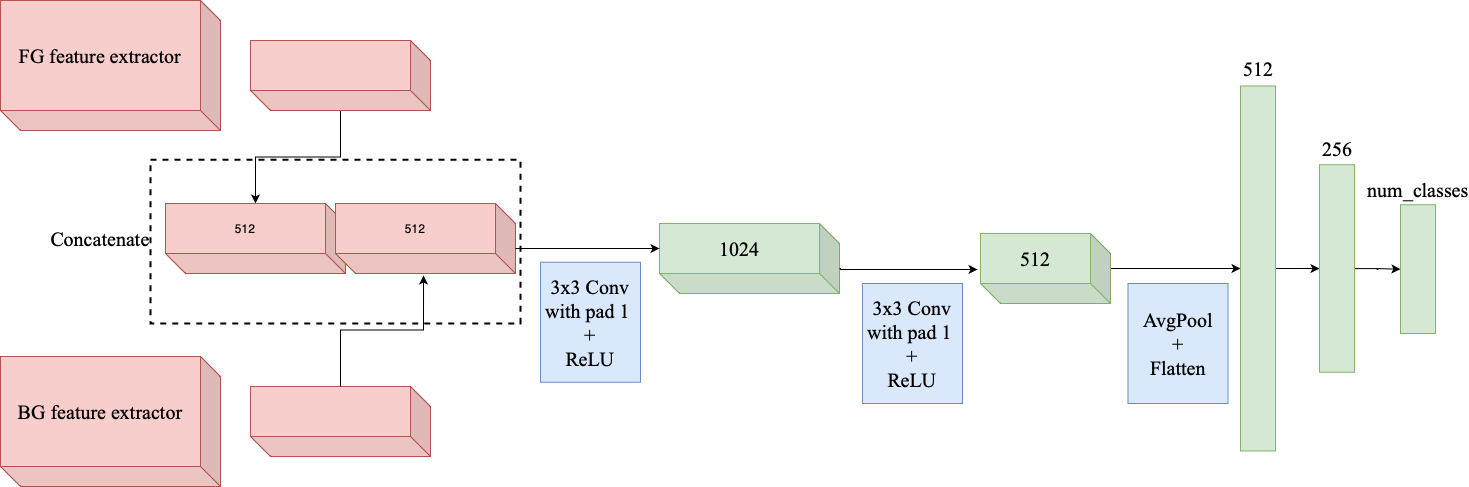}
    \caption{Cartoon of multimodal model. The red parts are frozen during training for classification task while the green ones are allowed to change.}
    \label{fig:cartoon}
\end{figure}

The model architecture is shown in Figure\ref{fig:cartoon}. We take the pre-trained foreground and background feature extractors and remove the classification head and the last convolutional layer. We then fuse background and foreground streams by concatenating the extracted features along the channel axis and further process it through two convolutional layers. The features at the end of this processing undergo average pooling and are then flattened and sent through a fully connected classifier head. 

% We compared performance among three image classifiers: one based on the foreground feature extractor with a fully connected layer at the end (unimodal foreground), the second one based on the background feature extractor with a fully connected layer at the end (unimodal background) and the third one based on the fusion of background and  foreground features as described above. The unimodal classifiers have the weights of the convolutional layers frozen and only the last fully connected layer is trained on the classification dataset. The joint model is trained similarly, with the weights of the pre-trained feature extractors being frozen but the fusion convolutional layers and the classification head is trained on the classification dataset. 

We compared three classifiers: unimodal foreground (foreground feature extractor + fully connected layer), unimodal background (background features extractor + fully connected layer), and a fusion (joint) model combining foreground and background features. The unimodal models have frozen convolutional weights with only the fully connected layer trained. The fusion model also has frozen feature extractors but the fusion convolutional layers and the classification head is trained on the classification dataset.

\subsection{Datasets}
We pruned the large MS COCO dataset to images with <=2 instances of a single object to minimize co-occurring context. Using bounding boxes, we constrained foreground to $\leq 50\%$ of images, where foreground was inside the box. This significantly reduced the dataset size but helped learn a representation with enough background information. Our final dataset had 24 classes with 7500 images, 75\% used for training and the remaining for testing.

\section{Results}

\subsection{Foreground and background feature extractors generate object and scene specific features}
\begin{figure}
    \begin{subfigure}{0.24\textwidth}
    \centering
    \includegraphics[width=0.7\linewidth]{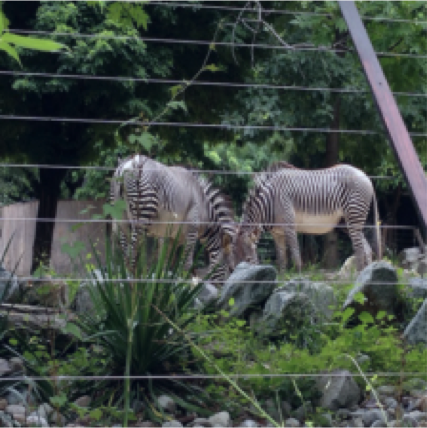}
    \caption{Original Image}
    \label{fig:zeb_img}
    \end{subfigure}
    \begin{subfigure}{0.24\textwidth}
    \centering
    \includegraphics[width=0.7\linewidth]{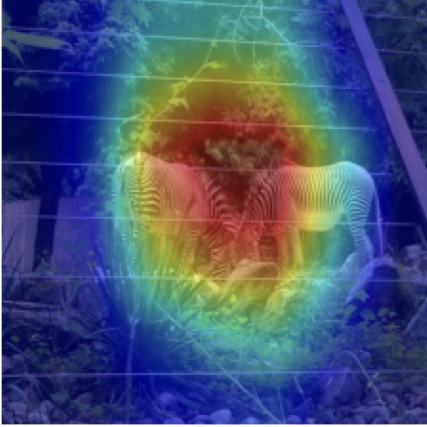}
    \caption{Foreground classifier}
    \label{fig:zeb_fg}
    \end{subfigure}
    \begin{subfigure}{0.24\textwidth}
    \centering
    \includegraphics[width=0.7\linewidth]{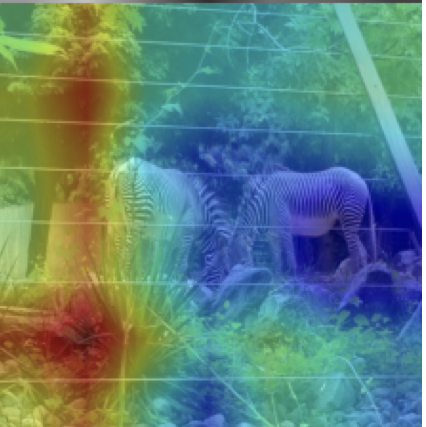}
    \caption{Background classifier}
    \label{fig:zeb_bg}
    \end{subfigure}
    \begin{subfigure}{0.24\textwidth}
    \centering
    \includegraphics[width=0.7\linewidth]{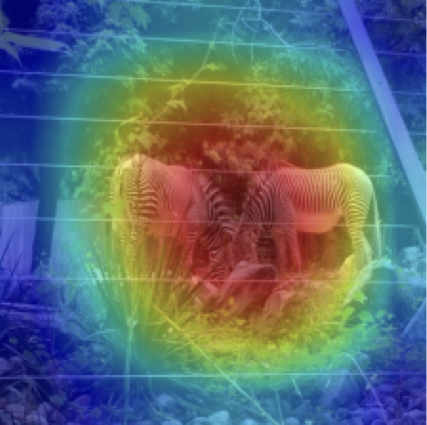}
    \caption{Joint classifier}
    \label{fig:zeb_joint}
    \end{subfigure}

    \begin{subfigure}{0.24\textwidth}
    \centering
    \includegraphics[width=0.7\linewidth]{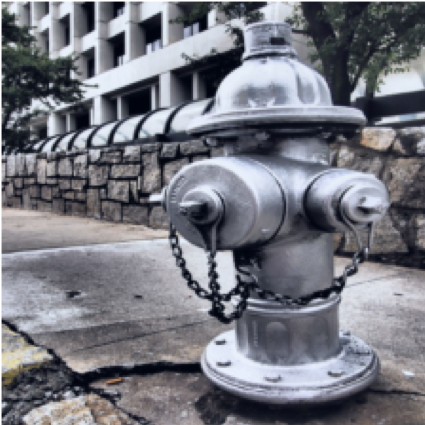}
    \caption{Original image}
    \label{fig:hyd_img}
    \end{subfigure}
    \begin{subfigure}{0.24\textwidth}
    \centering
    \includegraphics[width=0.7\linewidth]{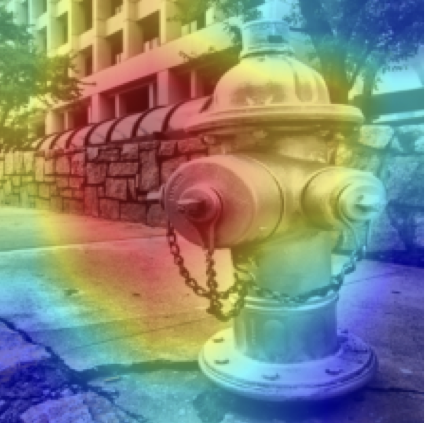}
    \caption{Foreground classifier}
    \label{fig:hyd_fg}
    \end{subfigure}
    \begin{subfigure}{0.24\textwidth}
    \centering
    \includegraphics[width=0.7\linewidth]{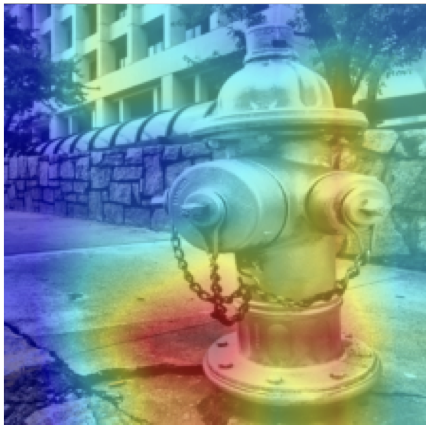}
    \caption{Background classifier}
    \label{fig:hyd_bg}
    \end{subfigure}
    \begin{subfigure}{0.24\textwidth}
    \centering
    \includegraphics[width=0.7\linewidth]{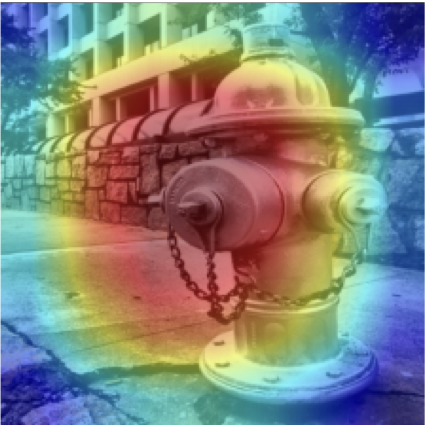}
    \caption{Joint classifier}
    \label{fig:hyd_joint}
    \end{subfigure}

\caption{Class activation maps for different images for the foreground, background and the joint classifiers}
\label{fig:GradCam}
\end{figure}

Class activation maps can provide us with information regarding the regions of the image that are most important for a particular classification task. \cite{cam}
We use GradCam tool \cite{gradcam} to visualise the regions identified by CNN to perform classification Figure \ref{fig:GradCam}). Specifically, the last convolutional layers of the foreground, background and joint network were used to calculate the Class Activation Maps. We found that the foreground and background networks attend to the object and scene, respectively, in most cases for MS COCO dataset, even though the CNNs themselves were trained on Imagenet and Places365. This indicates that the the foreground and background networks are extracting features related to the object and scene, respectively. The joint network attends more to the object since the convolutional layers are trained for classification in this case.

% Sentence about unexpected results

% FIGURE GRADCAM

\subsection{Blur affects foreground and background channels differently}
Below, we used two distinct CNNs with the same Resnet18 architecture to serve as the background and foreground feature extractors. The two different feature extractors represent the same images in different ways and have a different sensitivity to perturbations. 

\begin{figure}
    \begin{subfigure}{0.3\textwidth}
    \centering
    \includegraphics[width=0.6\linewidth]{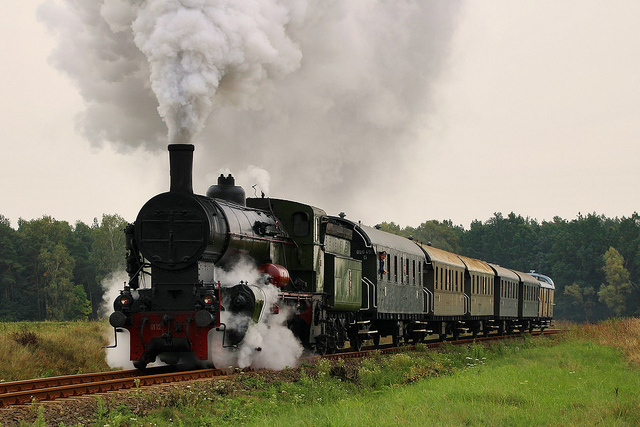}
    \caption{Normal image}
    \label{fig:normal_img}
    \end{subfigure}
    \begin{subfigure}{0.3\textwidth}
    \centering
    \includegraphics[width=0.6\linewidth]{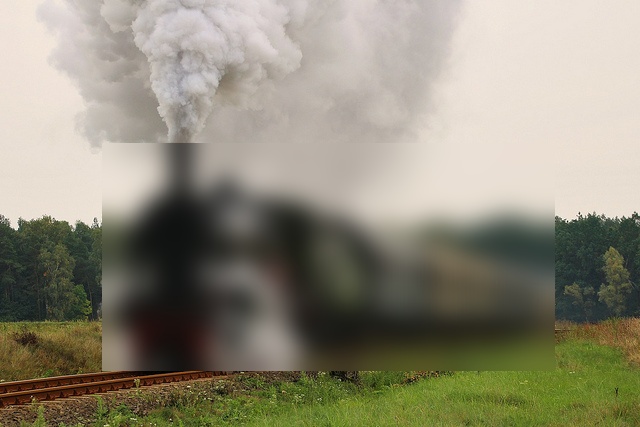}
    \caption{Only foreground blurred}
    \label{fig:blur_fg}
    \end{subfigure}
    \begin{subfigure}{0.3\textwidth}
    \centering
    \includegraphics[width=0.6\linewidth]{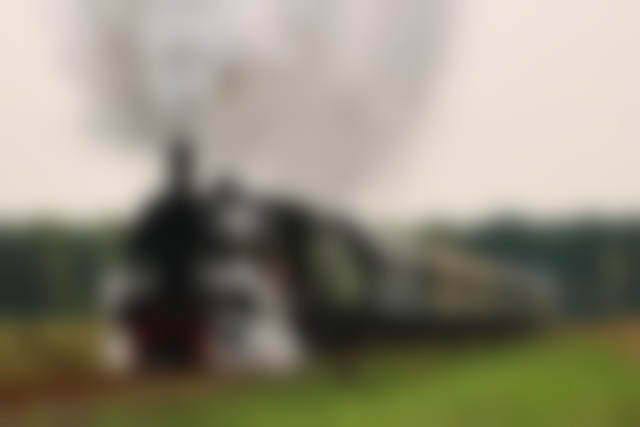}
    \caption{Entire image blurred}
    \label{fig:blur_whole}
    \end{subfigure}

\caption{Examples of normal and blurred images ($\sigma=12.0$) from the MSCOCO dataset.}
\label{fig:blur_imgs}
\end{figure}

Here is a shortened version:

Blur is a common real-world artifact and it is critical for an image classifier to maintain robustness to blur. We created blurred images by convolution of the foreground pixels within the bounding box with a Gaussian kernel. The blurred images were processed by the CNNs independently and features extracted from batch normalization layers.

\begin{figure}
  \centering
    \includegraphics[width=0.5\textwidth]{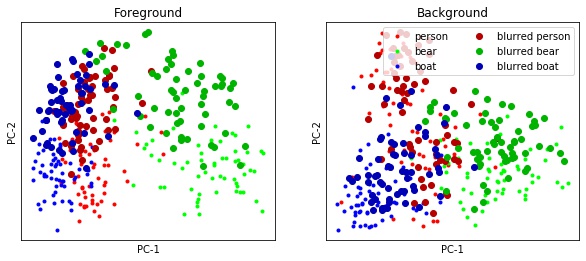}
    \caption{Effect of object blurring on background and foreground features. PCA projection to 2D-space is shown. Small and bright colored dots represent raw images while large and dark colored dots represent blurred images. (Left) entire subspace of the foreground features moved (up in this example) in presence of blur. (Right) after application of blur background features remain within the statistical subspace created by raw images. Each color represents a single image class.}
    \label{fig:PCA}
\end{figure}

To illustrate effect of blur on the high-level foreground and background features, we visualized the feature space using PCA to reduce dimensionality for few representative image classes. We found that blur had differing effects between the background (scene-centric) and foreground (object-centric) features. For the foreground features, blur caused a shift of the entire statistical representation subspace, i.e., all blurred images moved conjointly away from the non-blurred images (compare small vs large dots in Figure \ref{fig:PCA}, left). In contrast, impact of blur was much smaller for background features (Figure \ref{fig:PCA}, right). This shows, as one can expect, that blurring foreground (object) alone will have larger impact on the features extracted by the foreground feature extractor than the background one. This finding supports an idea that combined use of the foreground and background feature extractors in a multi-modal fashion may help to defend against distortions affecting the object.

\subsection{Effect of blurring the image on classification performance}

\begin{figure}
    \begin{subfigure}{0.5\textwidth}
    \includegraphics[width=0.8\linewidth]{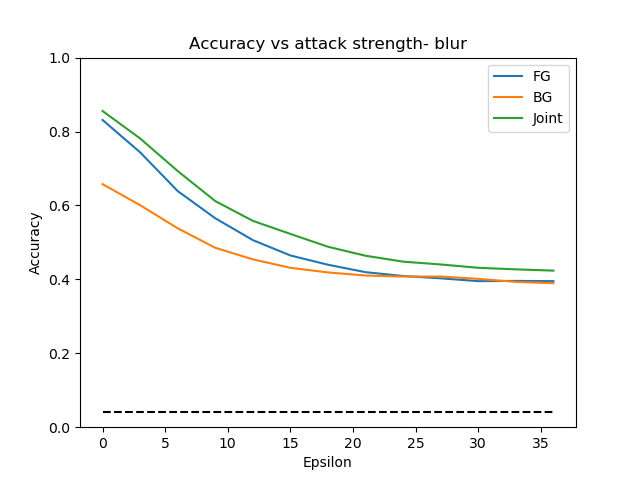}
    \caption{Test set 'All', only foreground blurred}
    \label{fig:all_fg}
    \end{subfigure}
    \begin{subfigure}{0.5\textwidth}
    \includegraphics[width=0.8\linewidth]{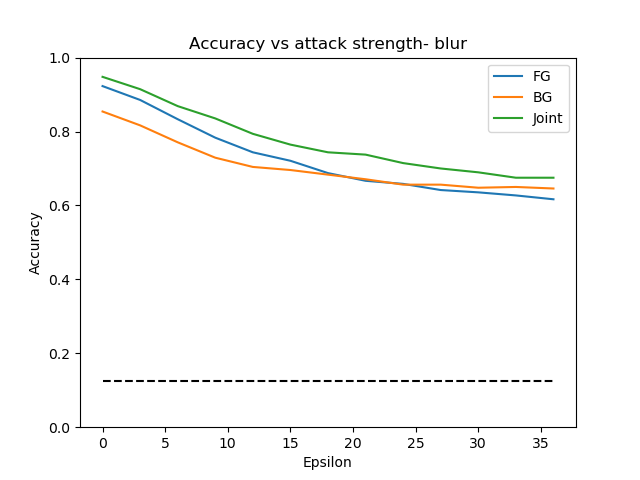}
    \caption{Test set 'Dissimilar', only foreground blurred}
    \label{fig:diss_fg}
    \end{subfigure}
    \begin{subfigure}{0.5\textwidth}
    \includegraphics[width=0.8\linewidth]{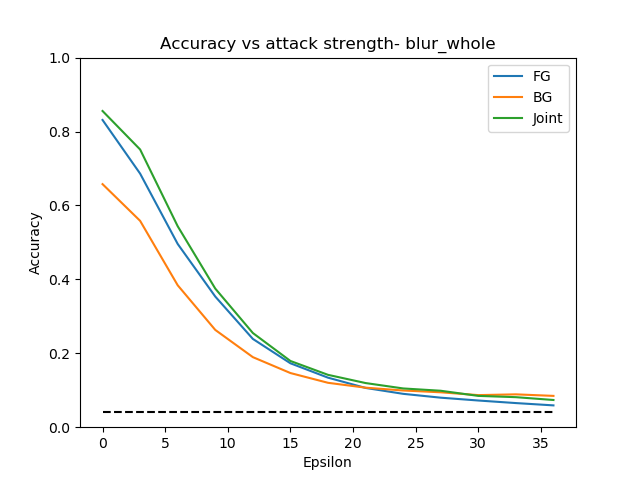}
    \caption{Test set 'All', whole image blurred}
    \label{fig:all_whole}
    \end{subfigure}
    \begin{subfigure}{0.5\textwidth}
    \includegraphics[width=0.8\linewidth]{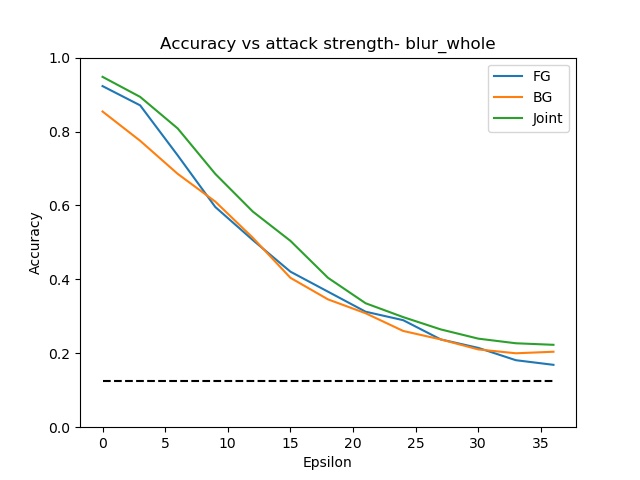}
    \caption{Test set 'Dissimilar', whole image blurred}
    \label{fig:diss_whole}
    \end{subfigure}
\caption{Effect of blurring images on classification performance. Panels a and b show the performance when only the foreground of the image is blurred as in Figure\ref{fig:blur_fg}, while c and d show the performance when the whole image is blurred as in Figure\ref{fig:blur_whole}. FG, BG and Joint denote the three different classifiers we have compared - foreground, background and the joint.}
\label{fig:blur_classif}
\end{figure}

% The scene or background of an image provides some context as to what the object in the image could be. For example, if the object is a horse, the background is more likely to be grass or a barn and is less likely to be the bedroom. This context provided by background may help in a classification task when the foreground of the image is perturbed. 
To test the hypothesis of increased robustness of the joint model, we performed a classification task on the images where either only the foreground pixels were blurred but the background pixels were intact or all the pixels were blurred (Figure \ref{fig:blur_classif}).
Our joint fusion method depends on associating context with object, thus we also tested separately images with Similar and Dissimilar background context. To do this, we used the "supercategories" given in the MS COCO dataset to refine test sets into two data sets: the "Dissimilar test set" where images come from distinct supercategories (e.g. "dog"-animal, "airplane"-vehicles, and "toilet"-indoor) and "All test set", where we did not make the distinction in terms of supercategories. 
%This changes our problem to either a coarse grained classification e.g. classes from different supercategories like “Animals”, “Vehicles”, “Indoor”, “Outdoor”, or a much harder fine grained classification, e.g. classes from the same supercategory of “Animals”. 

When only foreground was blurred, an increase in classification accuracy was evident for each subset of the MS COCO dataset (Similar or Dissimilar) and almost all values of $\sigma$ (Figure \ref{fig:all_fg} and Figure \ref{fig:diss_fg}). Specifically, if all images in the test set contained different contexts (e.g. some were indoor images and others were outdoor images), then the joint classier performed much better than the individual ones. This finding was expected because contexts in such case  contain  information specific to the underlying class. However, we found significant increases in classification with the joint network over the unimodal foreground classifier even when using all of the test data including those with similar and dissimilar background contexts.

% FIGURE SHOWING different classification curves

We further found that even when the entire image was blurred, the joint model performed better than unimodal ones (Figure \ref{fig:all_whole}, \ref{fig:diss_whole}). Specifically, the joint classifier performed better for the Dissimilar set, indicating that the context of background aids classification even when the background is blurred.

% \section{GradCam representations - something i guess}
\section{Conclusion}
We present a novel method for integrating context of an image using semantic data fusion to increase robustness to image blurring. By using features from distinct information streams, object-centered foreground and scene-centered background, we were able to maintain higher classification performance in the face of targeted blurring of the object in the image. We found that the degree of success for this method partially depends on the amount of variability in the background data. Thus, our method was more successful at maintaining robustness when objects came from distinct super categories of data with different and distinct backgrounds.

% \begin{ack}

% \end{ack}

%%%%%%%%%%%%%%%%%%%%%%%%%%%%%%%%%%%%%%%%%%%%%%%%%%%%%%%%%%%%
\bibliographystyle{unsrt}
\bibliography{neurips}

\begin{thebibliography}{1}

\bibitem{DBLP:journals/corr/BaltrusaitisAM17}
Tadas Baltrusaitis, Chaitanya Ahuja, and Louis{-}Philippe Morency.
\newblock Multimodal machine learning: {A} survey and taxonomy.
\newblock {\em CoRR}, abs/1705.09406, 2017.

\bibitem{gisiger_computational_2000}
Thomas Gisiger, Stanislas Dehaene, and Jean-Pierre Changeux.
\newblock Computational models of association cortex.
\newblock {\em Current Opinion in Neurobiology}, 10(2):250--259, April 2000.

\bibitem{pandya_association_1982}
Deepak~N. Pandya and Benjamin Seltzer.
\newblock Association areas of the cerebral cortex.
\newblock {\em Trends in Neurosciences}, 5:386--390, January 1982.
\newblock Publisher: Elsevier.

\bibitem{mars_412_2017}
R.~B. Mars, R.~E. Passingham, F.~X. Neubert, L.~Verhagen, and J.~Sallet.
\newblock 4.12 - {Evolutionary} {Specializations} of {Human} {Association} {Cortex}.
\newblock In Jon~H. Kaas, editor, {\em Evolution of {Nervous} {Systems} ({Second} {Edition})}, pages 185--205. Academic Press, Oxford, January 2017.

\bibitem{rosen_role_2017}
Maya~L. Rosen, Margaret~A. Sheridan, Kelly~A. Sambrook, Matthew~R. Peverill, Andrew~N. Meltzoff, and Katie~A. McLaughlin.
\newblock The {Role} of {Visual} {Association} {Cortex} in {Associative} {Memory} {Formation} across {Development}.
\newblock {\em Journal of Cognitive Neuroscience}, 30(3):365--380, October 2017.
\newblock Publisher: MIT Press.

\bibitem{zhou_learning_2014}
Bolei Zhou, Agata Lapedriza, Jianxiong Xiao, Antonio Torralba, and Aude Oliva.
\newblock Learning deep features for scene recognition using places database.
\newblock In {\em Advances in neural information processing systems}, pages 487--495, 2014.

\bibitem{imagenet}
Jia Deng, Wei Dong, Richard Socher, Li-Jia Li, Kai Li, and Li~Fei-Fei.
\newblock Imagenet: A large-scale hierarchical image database.
\newblock In {\em 2009 IEEE Conference on Computer Vision and Pattern Recognition}, pages 248--255, 2009.

\bibitem{cam}
Bolei Zhou, Aditya Khosla, Agata Lapedriza, Aude Oliva, and Antonio Torralba.
\newblock Learning deep features for discriminative localization.
\newblock In {\em 2016 IEEE Conference on Computer Vision and Pattern Recognition (CVPR)}, pages 2921--2929, 2016.

\bibitem{gradcam}
Ramprasaath~R. Selvaraju, Abhishek Das, Ramakrishna Vedantam, Michael Cogswell, Devi Parikh, and Dhruv Batra.
\newblock Grad-cam: Why did you say that? visual explanations from deep networks via gradient-based localization.
\newblock {\em CoRR}, abs/1610.02391, 2016.

\end{thebibliography}

\end{document}